%% file: main.tex
\documentclass[10pt,journal,compsoc]{IEEEtran_old}

\ifCLASSOPTIONcompsoc
  \usepackage[nocompress]{cite}
\else
  \usepackage{cite}
\fi

\usepackage{times}
\usepackage{epsfig}
\usepackage{graphicx}
\usepackage{amsmath}
\usepackage{amssymb}
\usepackage{subfigure}
\usepackage{hyperref}
\usepackage{seqsplit}
\usepackage{url}
\usepackage{algorithmic}
\usepackage{algorithm}
\usepackage{multirow}
\usepackage{threeparttable}
\usepackage{tablefootnote}
\usepackage{booktabs}
\usepackage{arydshln}
\usepackage{color}
\usepackage{bm}
\usepackage{subfiles}
\hypersetup{
    colorlinks=true,
    linkcolor=blue,
    filecolor=magenta,      
    urlcolor=magenta,
}
\def\etal{\emph{et al}. }
\def\ie{\emph{i.e.} }

\begin{document}

\title{VoxelTrack: Multi-Person $3$D Human Pose Estimation and Tracking in the Wild}

\author{Yifu Zhang,
        Chunyu Wang,
        Xinggang Wang,
        Wenyu Liu,
        Wenjun Zeng
\IEEEcompsocitemizethanks{\IEEEcompsocthanksitem Chunyu Wang and Wenjun Zeng are with Microsoft Research Asia.\protect\\
E-mail: \{chnuwa, wezeng\}@microsoft.com
\IEEEcompsocthanksitem Yifu Zhang, Xinggang Wang and Wenyu Liu are with Huazhong University of Science and Technology.\protect\\
Email: \{yifuzhang, xgwang, liuwy\}@hust.edu.cn}
}

\markboth{Journal of \LaTeX\ Class Files,~Vol.~14, No.~8, August~2015}%
{Shell \MakeLowercase{\textit{et al.}}: Bare Demo of IEEEtran.cls for Computer Society Journals}

\IEEEcompsoctitleabstractindextext{
\begin{abstract}
We present \emph{VoxelTrack} for multi-person $3$D pose estimation and tracking from a few cameras which are separated by wide baselines. It employs a multi-branch network to jointly estimate $3$D poses and re-identification (Re-ID) features for all people in the environment. In contrast to previous efforts which require to establish cross-view correspondence based on noisy $2$D pose estimates, it directly estimates and tracks $3$D poses from a $3$D voxel-based representation constructed from multi-view images. We first discretize the $3$D space by regular voxels and compute a feature vector for each voxel by averaging the body joint heatmaps that are inversely projected from all views. We estimate $3$D poses from the voxel representation by predicting whether each voxel contains a particular body joint. Similarly, a Re-ID feature is computed for each voxel which is used to track the estimated $3$D poses over time. The main advantage of the approach is that it avoids making any hard decisions based on individual images. The approach can robustly estimate and track $3$D poses even when people are severely occluded in some cameras. It outperforms the state-of-the-art methods by a large margin on three public datasets including Shelf, Campus and CMU Panoptic. 
\end{abstract}

}

\maketitle

\IEEEdisplaynotcompsoctitleabstractindextext

\IEEEpeerreviewmaketitle

\section{Introduction}
\subfile{1introduction}

\section{Related Work}
\subfile{2relatedwork}

\section{VoxelTrack: $3$D Pose Estimation}
\subfile{3voxeltrack1}

\section{VoxelTrack: $3$D Pose Tracking}
\subfile{3voxeltrack2}

\section{Datasets and Metrics}
\subfile{5datasets}

\section{Experiments}
\subfile{6experiments}

\section{Conclusion}
\subfile{7conclusion}

\ifCLASSOPTIONcompsoc
  \section*{Acknowledgments}
\else
  \section*{Acknowledgment}
\fi

This work was in part supported by NSFC (No. 61733007 and No. 61876212) and MSRA Collaborative Research Fund.

\ifCLASSOPTIONcaptionsoff
  \newpage
\fi

\bibliographystyle{IEEEtran}      
\bibliography{egbib}   

\begin{IEEEbiography}[{\includegraphics[width=1in,height=1.25in,clip,keepaspectratio]{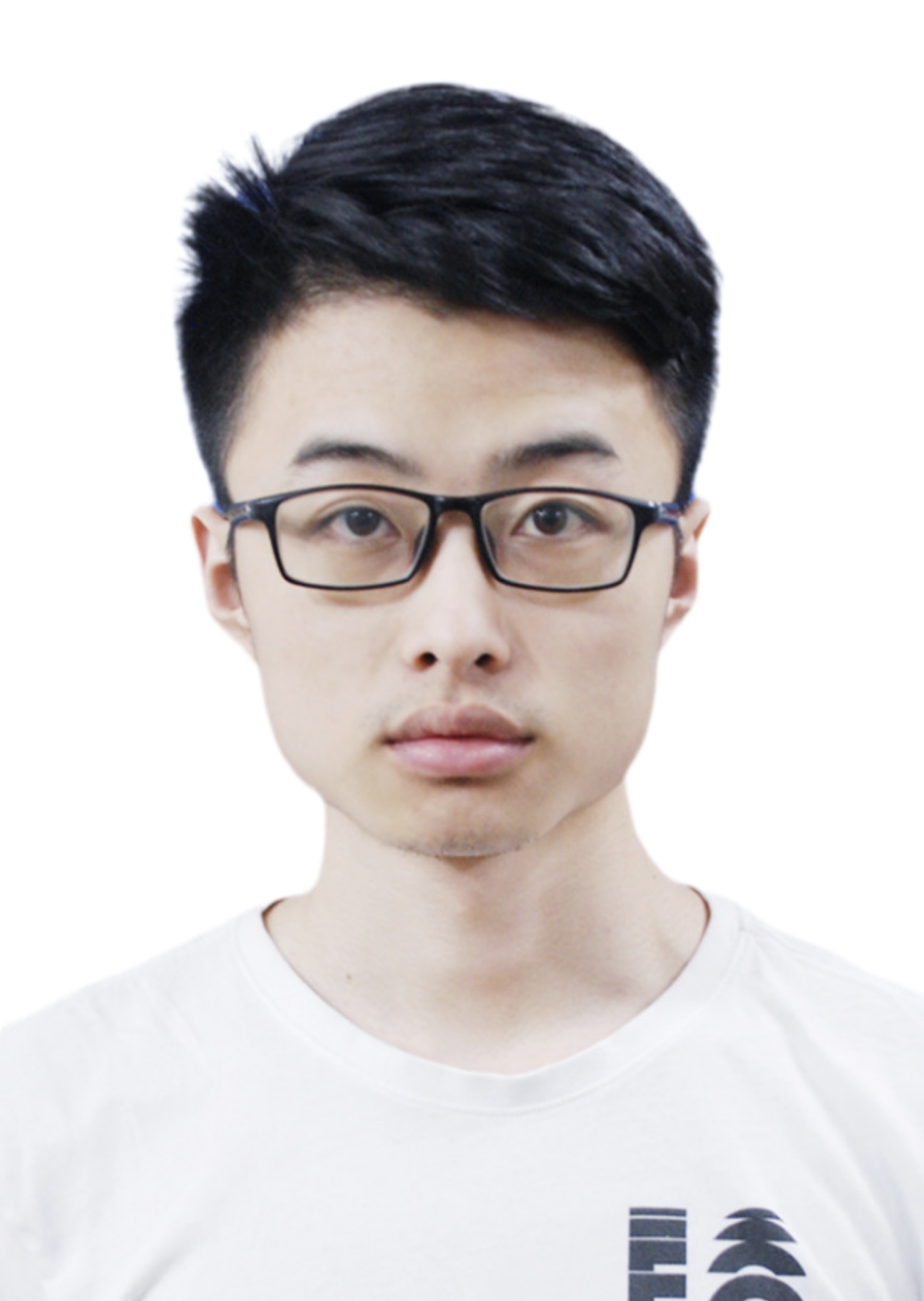}}] {Yifu Zhang} is a M.S. student in the School of Electronics Information and Communications, Huazhong University of Science and Technology (HUST), Wuhan, China. He received the B.S. degree from HUST in 2019. His research interests include computer vision and machine learning.
\end{IEEEbiography}

\begin{IEEEbiography}[{\includegraphics[width=1in,height=1.25in,clip,keepaspectratio]{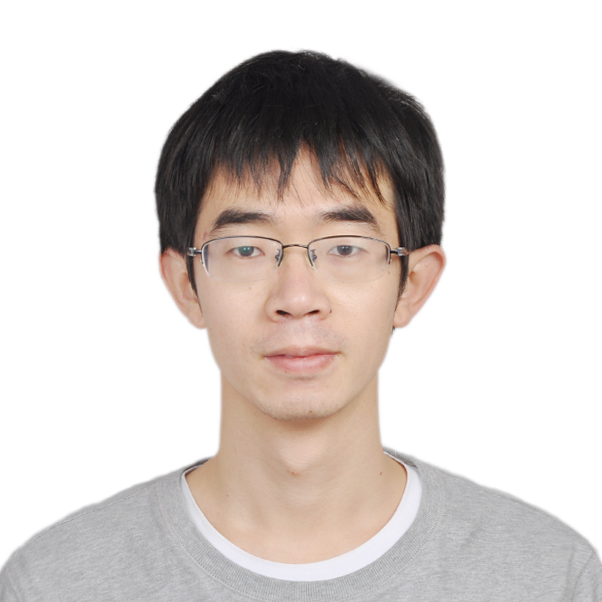}}]{Chunyu Wang} is a senior researcher in Microsoft Research Asia. He received his Ph.D in computer science from Peking University in 2016. His research interests include computer vision and machine learning algorithms, and their applications to solve real world problems.
\end{IEEEbiography}

\begin{IEEEbiography}[{\includegraphics[width=1in,height=1.25in,clip,keepaspectratio]{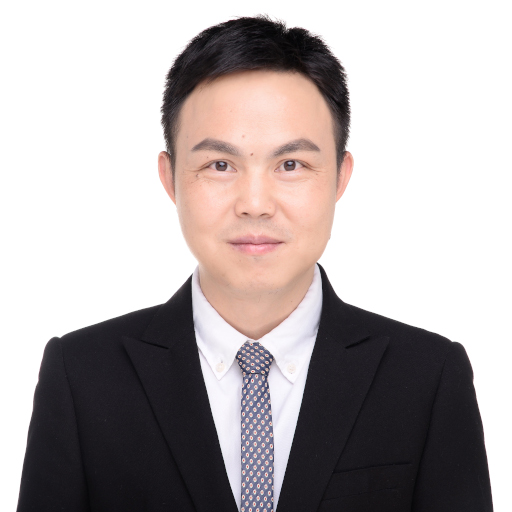}}] {Xinggang Wang (M’17)} received the B.S. and Ph.D. degrees in Electronics and Information Engineering from Huazhong University of Science and Technology (HUST), Wuhan, China, in 2009 and 2014, respectively. He is currently an Associate Professor with the School of Electronic Information and Communications, HUST. His research interests include computer vision and machine learning. He services as associate editors for Pattern Recognition and Image and Vision Computing journals and an editorial board member of Electronics journal.
\end{IEEEbiography}

\begin{IEEEbiography}[{\includegraphics[width=1in,height=1.25in,clip,keepaspectratio]{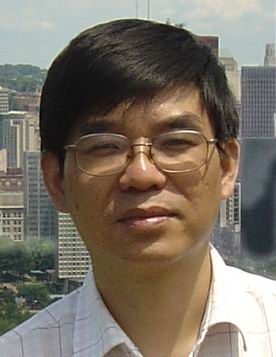}}] {Wenyu Liu (SM'15)} received the B.S. degree in Computer Science from Tsinghua University, Beijing, China, in 1986, and the M.S. and Ph.D. degrees, both in Electronics and Information Engineering, from Huazhong University of Science and Technology (HUST), Wuhan, China, in 1991 and 2001, respectively. He is now a professor and associate dean of the School of Electronic Information and Communications, HUST. His current research areas include computer vision, multimedia, and machine learning.
\end{IEEEbiography}

\begin{IEEEbiography}[{\includegraphics[width=1in,height=1.25in,clip,keepaspectratio]{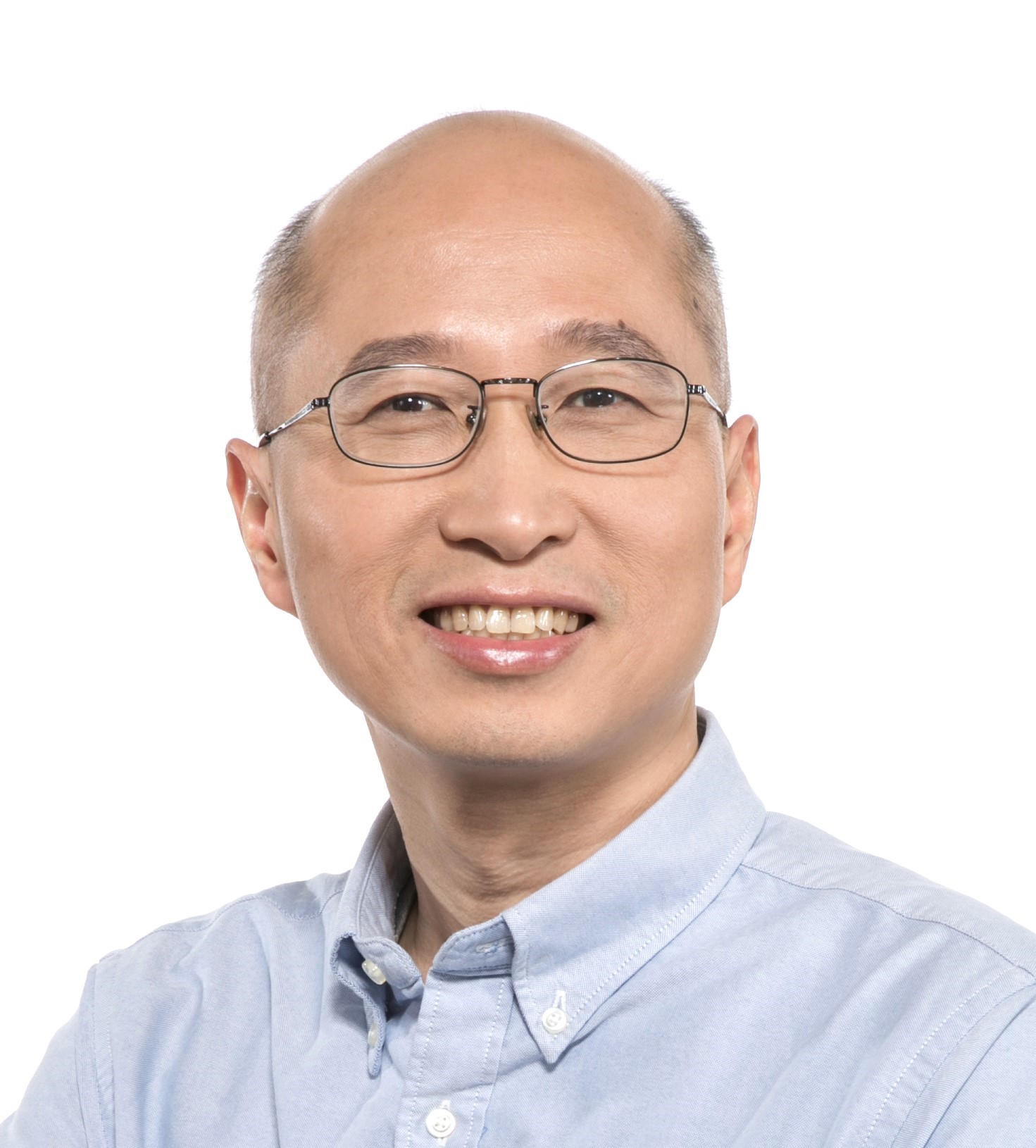}}] {Wenjun (Kevin) Zeng (M’97-SM’03-F’12)} is a Sr. Principal Research Manager and a member of the senior leadership team at Microsoft Research Asia. He has been leading the video analytics research empowering the Microsoft Cognitive Services, Azure Media Analytics Services, Office, Dynamics, and Windows Machine Learning since 2014. He was with Univ. of Missouri from 2003 to 2016, most recently as a Full Professor. Prior to that, he had worked for PacketVideo Corp., Sharp Labs of America, Bell Labs, and Panasonic Technology. Wenjun has contributed significantly to the development of international standards (ISO MPEG, JPEG2000, and OMA). He received his B.E., M.S., and Ph.D. degrees from Tsinghua Univ., the Univ. of Notre Dame, and Princeton Univ., respectively. His current research interests include mobile-cloud media computing, computer vision, and multimedia communications and security. He is on the Editorial Board of International Journal of Computer Vision. He was an Associate Editor-in-Chief of IEEE Multimedia Magazine, and was an AE of IEEE Trans. on Circuits \& Systems for Video Technology (TCSVT), IEEE Trans. on Info. Forensics \& Security, and IEEE Trans. on Multimedia (TMM). He was on the Steering Committee of IEEE Trans. on Mobile Computing and IEEE TMM. He served as the Steering Committee Chair of IEEE ICME in 2010 and 2011, and has served as the General Chair or TPC Chair for several IEEE conferences (e.g., ICME’2018, ICIP’2017). He was the recipient of several best paper awards. He is a Fellow of the IEEE.
\end{IEEEbiography}

\end{document}

%% file: 1introduction.tex
\IEEEPARstart{T}{}his work addresses the problem of multi-person $3$D pose estimation and tracking from a few cameras separated by wide baselines. The problem draws attention from multiple areas such as human pose estimation \cite{tu2020voxelpose,qiu2019cross,ma2021context,cao2017realtime,xiao2018simple,girdhar2018detect,qiu2019cross,wang2014robust}, person re-identification \cite{sun2018beyond,su2017pose,cheng2011custom} and multi-object tracking \cite{zhang2020fairmot,wang2019towards,baseline3dmot,long2018real,bergmann2019tracking,zhang2020fairmot,jin2017towards,zhu2017multi,doering2018joint,raaj2019efficient,henschel2019multiple,sun2019explicit,tanke2019iterative}. The problem can benefit many applications such as smart retail \cite{chen2020cross} and sport video analysis \cite{bridgeman2019multi}. 

\begin{figure}
	\centering
	\includegraphics[width=3.5in]{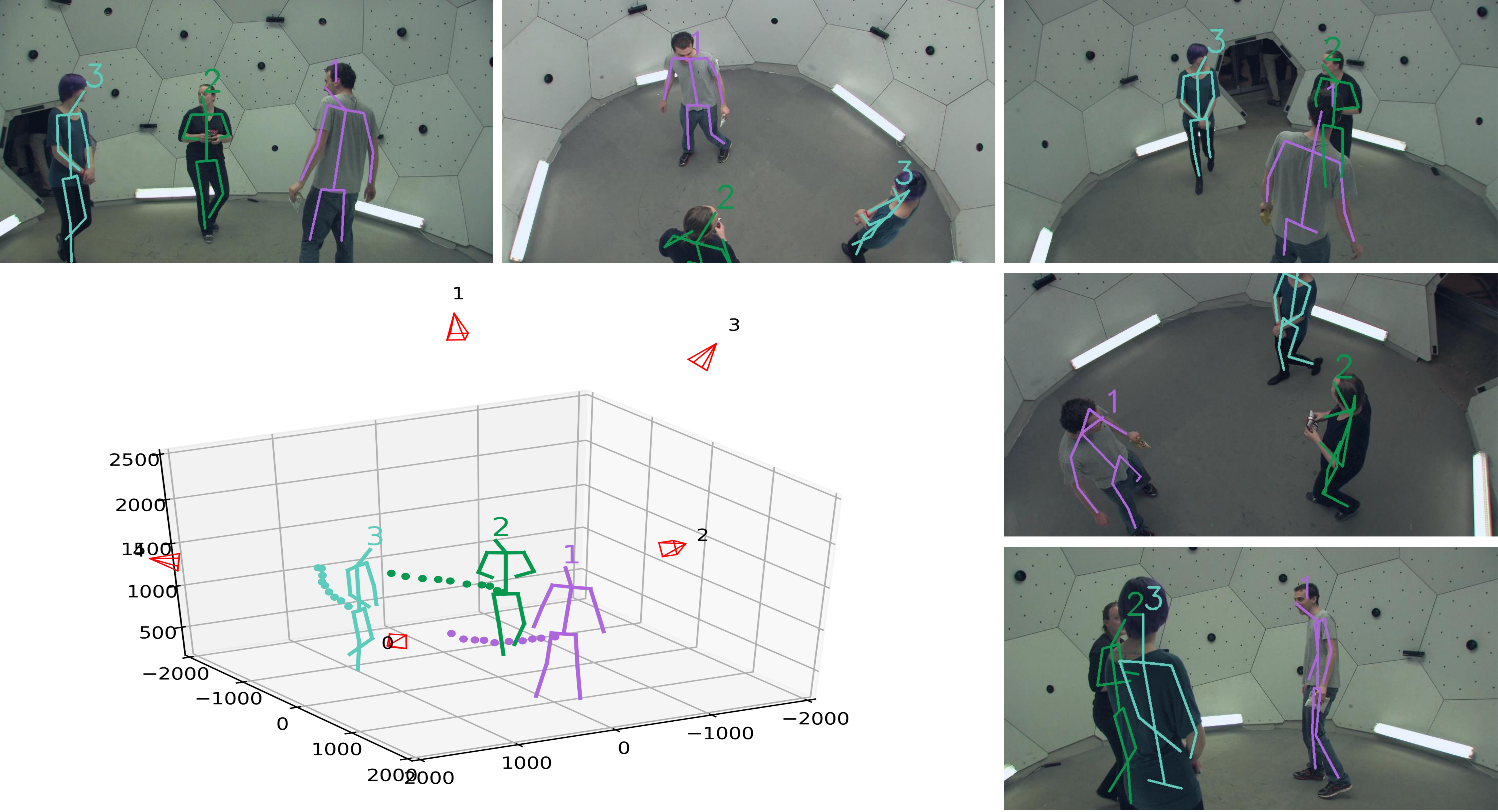}
	\caption{The top and right area show the images captured by five synchronized cameras. The bottom-left area shows the $3$D poses estimated by our approach. The numbers represent the identities of the estimates. The points with different colors represent the $3$D trajectories of the root joints of different persons over time. We project the estimated $3$D poses to images for visualization.  The red pyramids represent the locations and orientations of the five cameras.
	}
	\label{fig:teasing}
\end{figure}

Mainstream methods such as \cite{dong2019fast} usually address the problem by three separate models. First, they estimate $2$D poses in each camera view by CNN \cite{sun2019deep,cheng2020bottom,cao2017realtime}.  Second, they associate $2$D poses of the same person in different views based on epipolar geometry or image features, and recover the corresponding $3$D pose for each person by geometric methods \cite{hartley2003multiple,iskakov2019learnable,burenius20133d,qiu2019cross}. 
Third, they link the estimated $3$D poses over time by linear bipartite matching based on
keypoint locations and image features. The three tasks have been independently addressed by researchers from different areas which may unfortunately lead to degraded performance  (1) $2$D pose estimation is noisy especially when occlusion occurs; (2) $3$D estimation accuracy depends on the $2$D estimation and association results in all views; (3) unreliable appearance features caused by occlusion harms $3$D pose tracking accuracy. In the following, we discuss the challenges in detail and present an overview of how our end-to-end approach successfully addresses them.
\vspace{1em}

\subsection{Challenges in 2D Human Pose Estimation}
Introducing of CNN and large scale datasets \cite{andriluka14cvpr,lin2014microsoft,wu2017ai} has boosted $2$D pose estimation accuracy \cite{xie2020humble,cao2017realtime,li2019rethinking,sun2019deep,kreiss2019pifpaf,xiao2018simple,newell2016stacked,tompson2014joint} on benchmark datasets. However, even top-ranking methods suffer when we apply them to crowded scenes with severe occlusion and background clutter. Figure~\ref{fig:teasing} shows some images of this type from the Panoptic dataset \cite{Joo_2017_TPAMI}. On one hand, top-down pose estimators \cite{sun2019deep,chen2018cascaded}, which first detect all people in the image and then perform single person pose estimation for each detection, often fail to detect people that are mostly occluded. On the other hand, bottom-up methods \cite{cao2017realtime,newell2017associative,kreiss2019pifpaf}, which first detect all joints in an image and then group them into different instances, have limited capability to detect joints of small scale people. In summary, $2$D pose estimates are very noisy in real-world applications which will inevitably cause negative and irreversible impact to the $3$D pose estimation step.

\subsection{Challenges in 3D Human Pose Estimation}
To estimate $3$D poses of multiple people, we first need to associate the $2$D poses of the same person in different views.
This can be achieved by matching $2$D poses based on epipolar geometry. Some methods \cite{dong2019fast} also use image features to improve robustness. But still, it is a challenging task because $2$D pose estimates may have large errors and appearance features may be corrupted when occlusion occurs. For example in Figure~\ref{fig:teasing}, we can see that some people are only partially visible in some camera views. So features of the people in those camera views are very different from those in normal views. After we obtain the corresponding joint locations of each person, we estimate their $3$D locations by triangulation \cite{hartley2003multiple,iskakov2019learnable} or pictorial structure models \cite{qiu2019cross,amin2013multi}. The final $3$D pose estimation accuracy largely depends on the cross view association step which in turn depends on the accuracy of the $2$D pose estimation step.

\subsection{Challenges in 3D Human Pose Tracking}
The third step is to link the estimated $3$D poses over time and obtain a number of tracks. Although the task itself is rarely studied, there are many works on $2$D pose tracking in videos \cite{xiao2018simple,iqbal2017posetrack,insafutdinov2017arttrack,girdhar2018detect,xiu2018poseflow,ning2019lighttrack} which can be easily extended to track $3$D poses. The fundamental problem in the task is to measure the similarity between every pair of poses between neighboring frames and then solve the classic assignment problem by bipartite matching \cite{kuhn1955hungarian}. Two sources of information have been used to compute pose similarity. The first is based on motion cues which uses Kalman Filter or optical flow to predict future positions of the tracklets and prevents them from being linked to the poses (in the current frame) which are far from the predictions. This effectively promotes smoothness in tracking results. The second class of information is appearance features computed from images. However, when occlusion occurs, appearance features may be corrupted and unreliable.

\begin{figure*}
	\centering
	\includegraphics[width=6in]{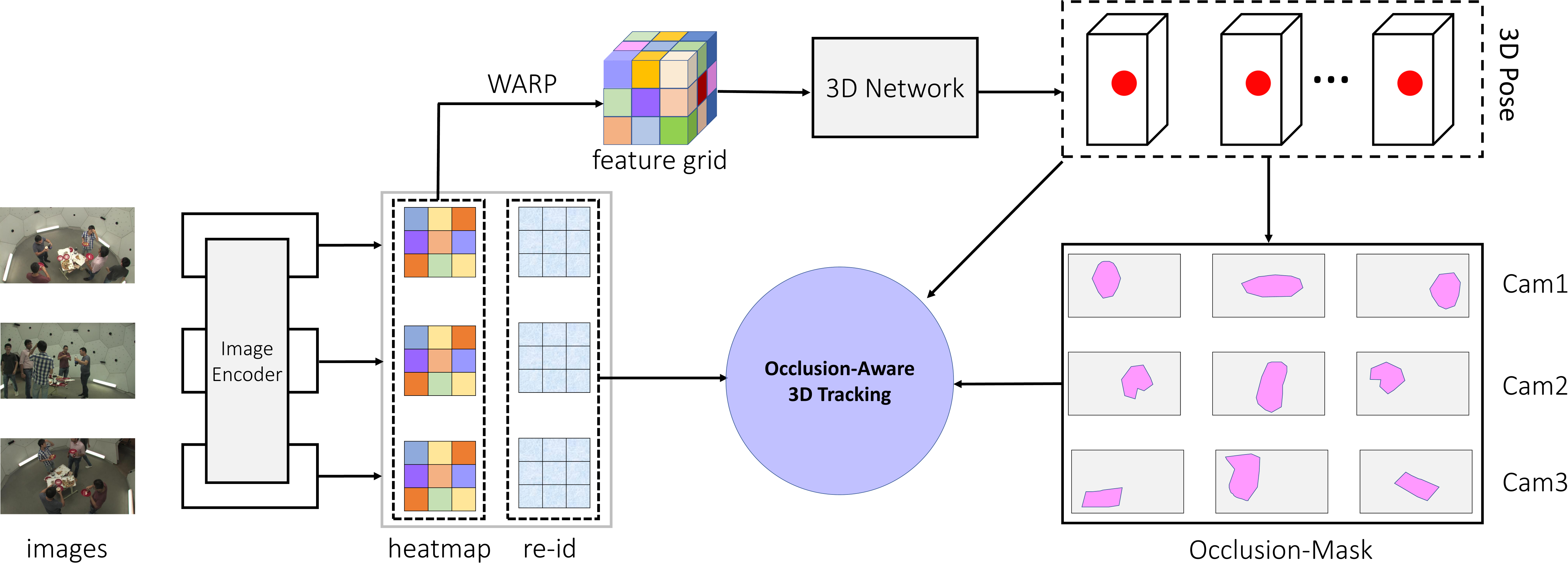}
	\caption{Overview of \emph{VoxelTrack} for $3$D pose tracking. Given multi-view images as input, it first estimates pixel-wise pose heatmaps and Re-ID features for each view. Then the heatmaps are warped to construct a $3$D feature grid which is fed to a $3$D pose network to estimate $3$D poses. We estimate person-person occlusion relationships in each view by using the $3$D poses and camera parameters. Finally, we perform $3$D pose tracking with the $3$D poses, occlusion-masks and Re-ID features as input.
	}
	\label{fig:pipeline}
\end{figure*}

\subsection{Overview of Our Approach}
We present an approach termed \emph{VoxelTrack} for robustly tracking $3$D poses of multiple people in challenging environment. \textbf{This builds on, and gives a more detailed description of our preliminary work \emph{VoxelPose} \cite{tu2020voxelpose} which estimates $3$D poses from multiple cameras.} Our new contributions include 1) extending the work to be able to track $3$D poses over time and 2) exploiting sparseness of the $3$D representation to improve the inference speed so that it can be applied to very large 3D spaces such as football court. An overview of our approach is shown in Figure \ref{fig:pipeline}.

\vspace{1em}
\noindent
\textbf{(I) $2$D Feature Extraction}
Given synchronized images as input, it first estimates $2$D pose heatmaps \cite{cheng2020bottom} and Re-ID feature maps \cite{zhang2020simple} by a CNN-based image encoder. Recall that a pose heatmap encodes per-pixel likelihood of a joint. Similarly, a Re-ID feature at each pixel encodes identity  embedding of the person centered at the pixel. Note that we do not recover $2$D poses from the heatmaps because they are usually very noisy. Instead, we keep the ambiguity in the representation and postpone decision making to the later $3$D stage in which multi-view information is available to resolve the ambiguity. 
\vspace{0.5em}

\noindent
\textbf{(II) Holistic $3$D Representation}
To avoid making incorrect decisions in each view, our approach directly operates in the $3$D space by fusing information from all views. Specifically, we divide the $3$D motion capture space by regular voxels and compute a feature vector for each voxel by inversely projecting the $2$D heatmap vector at the corresponding location in each view using camera parameters. The resulting $3$D heatmap volume, which carries positional information of body joints, will be fed to $3$D network to estimate the likelihood of each voxel having a particular body joint. The holistic $3$D representation elegantly avoids cross view association of $2$D poses. In parallel, we also compute a $3$D Re-ID feature volume for tracking $3$D poses over time.
\vspace{0.5em}

\noindent
\textbf{(III) $3$D Pose Estimation} The resulting $3$D heatmap volume is sparse. So we apply a lightweight network with sparse $3$D CNNs \cite{yan2018second} to the volume to estimate a $3$D heatmap which encodes per-voxel likelihood of all joints. We first obtain a number of person instances by finding peak heatmap values of the root joint. Then for each instance, we crop a smaller fixed-size region from the volume centered at the root joint and use an independent lightweight network to estimate $3$D heatmaps of all joints that belong to the instance. Finally, we obtain the $3$D pose of the instance by computing expectation over the heatmaps.
\vspace{0.5em}

\noindent
\textbf{(IV) $3$D Pose Linking}
Considering that occlusion occurs frequently in real-world applications, we introduce an occlusion-aware multi-view feature fusion strategy when we compute the Re-ID feature volume. The idea is that, for each $3$D pose, we estimate whether it is occluded by other people in each camera view which determines whether the features in this view will be used for fusion. We link the $3$D poses over time by bipartite matching based on the fused features and the $3$D locations. The occlusion-aware matching strategy eliminates the harm of unreliable appearance features.
\vspace{0.5em}

\noindent
\textbf{How VoxelTrack Addresses the Challenges?}

The most prominent advantage of \emph{VoxelTrack} is that it does not require to do $2$D pose estimation in every camera view nor pose association in different views as in previous works which is error-prone. Instead, all hard decisions are postponed and made in the $3$D space after fusing the inversely projected $2$D image features from all views. As a result, the ``end-to-end'' inference style effectively avoids error accumulation. In addition, the representation is robust to occlusion because it fuses the features in all camera views (a joint occluded in one view may be visible in other views). 

We evaluate our approach on three public datasets including the CMU Panoptic \cite{Joo_2017_TPAMI}, Shelf \cite{belagiannis20153d} and  Campus \cite{belagiannis20153d} datasets. \emph{VoxelTrack} outperforms the existing methods by a large margin which validates the advantages of performing tracking in $3$D space. More importantly, the estimation and tracking results are very stable even when severe occlusion occurs in some camera views. We also evaluate different factors in our approach such as occlusion-masks, similarity metrics and network structures. In addition, we find that the 3D network can be accurately trained on automatically generated synthetic heatmaps. This is possible mainly because the heatmap based $3$D feature volume representation is a high level abstraction that is disentangled from appearance/lighting, etc. This favorable property dramatically enhances the applicability of the approach. The whole system runs at 15 FPS with 5 camera views as input on a single 2080Ti GPU.

%% file: 2relatedwork.tex
In this section, we briefly review the existing work which are related to $3$D pose tracking including $2$D pose estimation, $3$D pose estimation, box-level human tracking and key-point-level human tracking.

\subsection{2D Human Pose Estimation}
Estimating $2$D poses in images has been a long-standing goal in computer vision \cite{yang2011articulated,chen2014articulated,andriluka2009pictorial,ferrari2008progressive}. Before deep learning, this had often been approached by modeling human as a graph and estimating the locations of the graph nodes in images according to image features and structural priors. Development of CNNs \cite{tompson2014joint,cao2017realtime,newell2016stacked,xiao2018simple} has led to remarkable accuracy improvement on benchmark datasets. In particular, introducing of the heatmap representation \cite{tompson2014joint}, which encodes per-pixel likelihood of body joints for each image, has dramatically improved the robustness of pose estimation. In fact, this probabilistic representation has become the de facto standard for pose estimation and is the main factor behind the success of those approaches. Our approach also uses this representation. But different from \cite{tompson2014joint}, we do not make hard decisions on heatmaps to recover joint locations because they are unreliable when occlusion occurs.
\vspace{1em}

When an image has multiple people, an additional challenge is to group the detected joints into different instances. Existing multi-person pose estimation methods can be classified into two classes based on how they do grouping: top-down and bottom-up approaches. Top-down approaches \cite{xiao2018simple,he2017mask,papandreou2017towards,li2019rethinking}
operate in two steps: detecting all people in the image by a number of boxes and then performing single person pose estimation for each box. They crop the image according to the boxes and normalize the image patches to have the same scale which notably improves the estimation accuracy. However, those approaches suffer a lot when a large part of a person is occluded because detection in such cases barely gets satisfying results. In addition, it suffers from the scalability issue because the computation time increases linearly as the number of people in images.

In contrast, bottom-up approaches \cite{cao2017realtime,xiao2018simple,newell2017associative,kreiss2019pifpaf}
first detect all joints in the image, and then group them into instances according to spatial and appearance affinities among them. However, since scales of different instances in a single image may vary significantly, pose estimation results are generally worse than those of the top-down methods. But joint detection is more robust to occlusion. The $2$D pose estimation module in \emph{VoxelTrack} is a bottom-up approach. However, we do not group joints into instances. Instead, we keep the heatmap representation and warps heatmaps of all views to a common $3$D space in order to detect $3$D person instances. So joint association  is implicitly accomplished.

\subsection{3D Human Pose Estimation}

There are two challenges in multi-person $3$D pose estimation. First, it needs to associate joints of the same person as discussed previously. Second, it needs to associate the $2$D poses of the same person in different views based on appearance features \cite{dong2019fast,bridgeman2019multi} or geometric features \cite{chen2020cross} which is unstable when people are occluded. Some methods adopt model-based methods by maximizing the consistency between the model projections and image observations. For example, the pictorial structure model is extended to deal with multiple person $3$D pose estimation in \cite{belagiannis20143d,belagiannis20153d}. However, the interactions across people introduce loops to the graph which notably complicates optimization. These challenges limit the $3$D pose estimation accuracy of those methods.

\vspace{0.5em}

Dong \etal \cite{dong2019fast} propose a multi-way matching algorithm to find
cycle-consistent correspondences of detected 2D poses across multiple views using both appearance and geometric cues, which is able to prune false detections and deal with partial overlaps between views. Then they recover the $3$D pose for each person using triangulation-based methods. Chen \etal \cite{chen2020cross} exploit temporal consistency in videos to match the detected 2D poses with the estimated 3D poses directly in $3$D space and update the $3$D poses iteratively via the
cross-view multi-human tracking. This novel formulation improves both accuracy and efficiency. However, both approaches are venerable to inaccurate $2$D pose estimates in each view which is often the case in practice.

\vspace{0.5em}

Our work is better than the pictorial structure models \cite{belagiannis20143d,belagiannis20153d} because it does not suffer from local optimum and does not need the number of people in each frame to be known as input. It differs from the model-free methods \cite{dong2019fast,bridgeman2019multi,chen2020cross} in that it elegantly avoids the two association problems. The approach is readily applicable to large spaces such as basketball court. The computation time is hardly affected by the number of people in the environment because we use the coarse-to-fine approach to divide space into voxels and the lightweight sparse $3$D CNN.

\subsection{Human Pose Tracking}

Human pose tracking is related to box-level object tracking which aims to estimate trajectories of objects in videos. Most methods follow the \emph{tracking-by-detection} paradigm, which first detect objects in each frame and then link them over time. The key is to compute similarity between detections and tracklets. One class of methods \cite{bewley2016simple,bochinski2017high,zhu2018online,bergmann2019tracking} use location and motion cues. For example, Kalman Filter \cite{kalman1960new} or optical flow are used to predict future tracklet positions and then they compute the distance between the predicted and detected object positions as the similarity. Another class of methods \cite{wojke2017simple,long2018real,wang2019towards,zhang2020fairmot} use image features to compute similarity. The first class are fast and effective for short-range linking while the second are better at handling long-range linking which is critical to track objects that re-appear after being occluded for a while. To track objects in multiple cameras, some works
\cite{ristani2018features,maksai2017non,liu2017multi,yoon2018multiple} propose to first detect $2$D boxes in each view and then link them across both time and view points according to appearance similarity. Some other works use multi-view geometry and location cues to track ground plane detections \cite{fleuret2007multicamera,peng2015robust,chavdarova2017deep,baque2017deep,chavdarova2018wildtrack}, $3$D points \cite{liu2016multi} or 3D centroid-with-extent detections \cite{ong2020bayesian}. 

Compared to box-level tracking, pose tracking has access to finer-grained joint locations. Some offline trackers such as \cite{iqbal2017posetrack,insafutdinov2017arttrack,keuper2015efficient} formulate pose tracking as a graph partitioning problem in which the joints of the same person in different frames are expected to be connected while the joints of different persons are disconnected. Some online trackers \cite{pirsiavash2011globally,girdhar2018detect,xiao2018simple,xiu2018poseflow,raaj2019efficient,ning2020lighttrack} solve the problem by bipartite matching which first estimate $2$D poses in the current frame and then link them to the closest tracklets, respectively. In \cite{xiao2018simple}, optical flow is used as the motion model to reduce missing detections and the similarity is computed by the human joint distances. In a recent work \cite{ning2020lighttrack}, a Graph Convolution Network Re-ID model is used to extract Re-ID features based on all joints instead of bounding boxes.

To our best, few works have systematically studied $3$D pose tracking in multiple cameras which is the focus of this work. Different from previous multi-view box-level tracking methods, we have more precise $3$D keypoint coordinates which allows us to reliably reason about occlusion. We combine keypoint distances and occlusion-aware appearance feature distances to compute similarity and achieve very stable tracking results in challenging scenes.

\begin{figure*}
	\centering
	\includegraphics[width=5in]{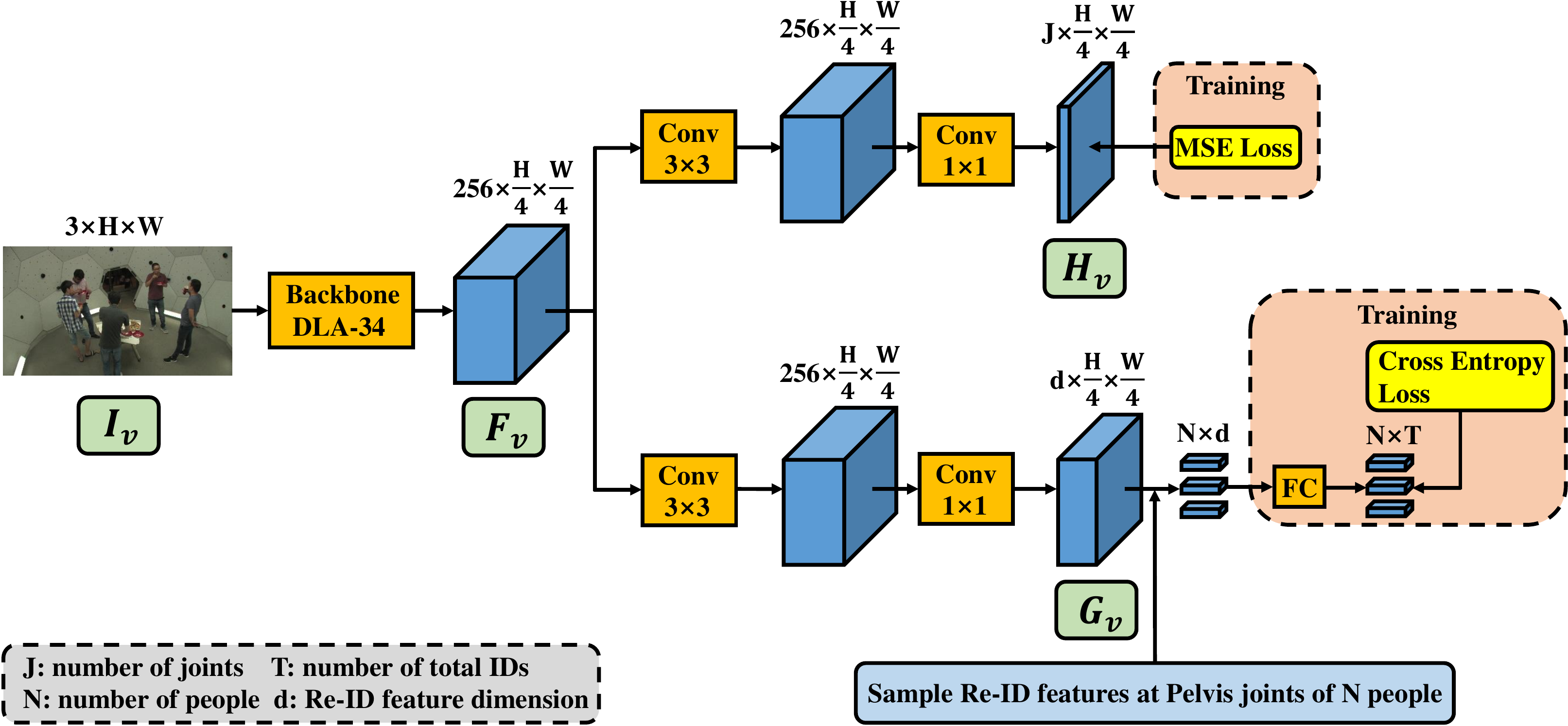}
	\caption{The network structure for estimating $2$D pose heatmaps and Re-ID features. $\mathbf{I}_v$, $\mathbf{F}_v$, $\mathbf{H}_v$ and $\mathbf{G}_v$ represent the image, backbone feature map, $2$D heatmap and Re-ID feature map of camera $v$, respectively. 
	}
	\label{fig:voxeltrack_2d}
\end{figure*}

%% file: 3voxeltrack1.tex
In this section, we present the first part of \emph{VoxelTrack} which estimates $3$D poses of all people in the environment. This includes estimating $2$D pose heatmaps, Re-ID features and $3$D poses from multiview images.

\subsection{Backbone Network}
As shown in Figure \ref{fig:voxeltrack_2d}, we use DLA-34 \cite{zhou2019objects} as backbone network to extract intermediate features for images of all views independently. The DLA network \cite{yu2018deep} was first proposed for image classification. We use a recent variant \cite{zhou2019objects} for dense prediction tasks which uses iterative deep aggregation to increase feature map resolution. It takes an image $\mathbf{I}_v \in \mathcal{R}^{3 \times H \times W}$ from view $v$ as input and outputs a feature map $\mathbf{F}_v \in \mathcal{R}^{C \times \frac{H}{4} \times \frac{W}{4}}$. The feature map will be fed to two networks to estimate $2$D pose heatmaps and Re-ID features, respectively as will be discussed subsequently.

\subsection{$2$D Pose Heatmap Estimation}
We use a simple network to estimate $2$D pose heatmaps $\mathbf{H}_v \in \mathbf{R}^{J \times \frac{H}{4} \times \frac{W}{4}}$ from backbone features $\mathbf{F}_v$ where $J$ is the number of body joint types. The network consists of two convolutional layers as shown in Figure \ref{fig:voxeltrack_2d}. A heatmap encodes per-pixel likelihood of a body joint which is a common surrogate representation for human pose used in many work \cite{cao2017realtime,xiao2018simple}. We train the $2$D pose heatmaps by minimizing:
\begin{equation}
    \mathcal{L_{\text{2D}}}={\|\mathbf{H}_{v}^{*} - \mathbf{H}_v\|_2}, 
\end{equation}
where $\mathbf{H}_v^* \in \mathbf{R}^{J \times \frac{H}{4} \times \frac{W}{4}}$ is the ground truth $2$D pose heatmaps computed following \cite{xiao2018simple}. Different from the previous works, we do not make any hard decisions on $2$D heatmaps but use them as input to our $3$D pose network.

\subsection{Re-ID Features}
We use a simple network to estimate Re-ID feature maps $\mathbf{G}_v \in \mathcal{R}^{d \times \frac{H}{4} \times \frac{W}{4} }$ from backbone features $\mathbf{F}_v$ where $d$ is the dimension of Re-ID features. It consists of two convolutional layers as shown in Figure \ref{fig:voxeltrack_2d}. Inspired by FairMOT \cite{zhang2020fairmot}, for each person in an image, we sample a $d$-dimensional feature from the feature maps at the pelvis joint as its Re-ID features. We use a Fully Connected (FC) network and a softmax operation to map the sampled features to a one-hot vector $\mathbf{P} = \{\mathbf{p}(t), t \in [1,T] \}$ representing the person's identity. Denote the one-hot representation of the GT class label as $\mathbf{L}^i{(p)}$. We train the Re-ID network as a classification task using cross entropy loss as follows:
\begin{equation}
\label{eq:id}
    \mathcal{L_{\text{ID}}} = - \sum_{i=1}^{N} \sum_{t=1}^{T} \mathbf{L}^i{(t)} \text{log}(\mathbf{p}(t)),
\end{equation}
where $N$ is the number of people in the image and $T$ is the total number of unique people in the training dataset. During training, we use ground-truth pelvis joint locations in images to extract Re-ID features. During testing, we project estimated $3$D pelvis locations to images to sample Re-ID features. 
During testing, we use the sampled Re-ID features before the classification layer to represent each person as will be described in the following sections. 

\begin{figure*}
	\centering
	\includegraphics[width=6in]{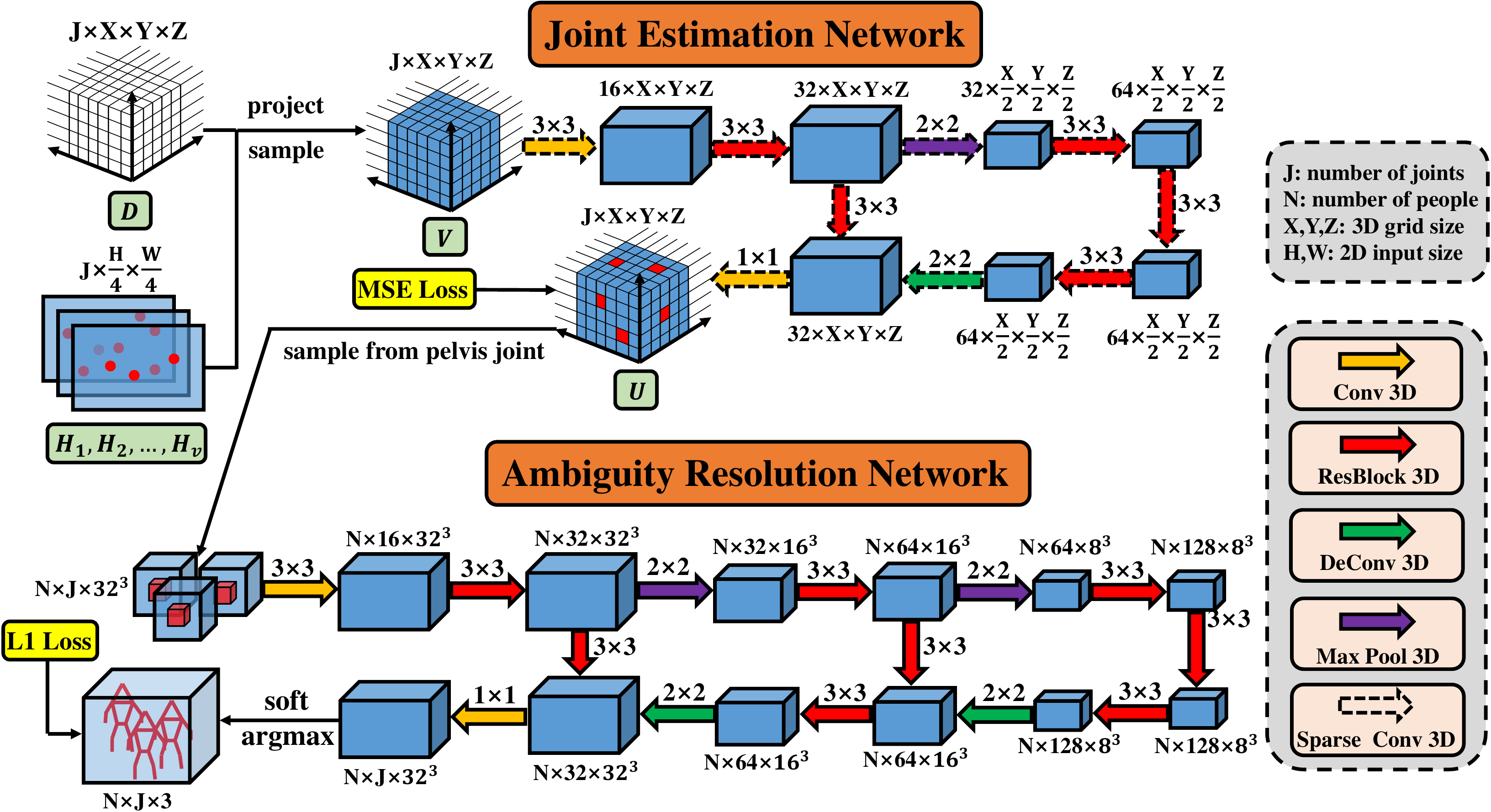}
	\caption{The network structure for estimating $3$D joint heatmaps and $3$D poses. We first use a Joint Estimation Network (JEN) to get the $3$D heatmaps of all the joints and then use a Ambiguity Resolution Network (ARN) to get the $3$D poses. $\mathbf{H}_v$, $\mathbf{D}$, $\mathbf{V}$, $\mathbf{U}$ represent the $2$D heatmap, empty discrete voxels, feature vectors of the voxels and the $3$D heatmap, respectively. }
	\label{fig:jointnetwork}
\end{figure*}

\subsection{$3$D Joint Estimation}

We discretize the $3$D motion space by $X \times Y \times Z$ discrete voxels $\{\mathbf{D}^{x,y,z} \}$. Each voxel is a candidate location for body joints. In order to reduce quantization error, we usually set the size of a voxel to be as small as possible ($62.5$mm in this paper). We compute a feature vector for each voxel by calculating average $2$D heatmap values sampled at its projected locations in all camera views. Denote the $2$D heatmap of view $v$ as $\mathbf{H}_v \in \mathcal{R}^{J \times \frac{H}{4} \times \frac{W}{4}}$ where $J$ is the number of body joints. For each voxel $\mathbf{D}^{x,y,z}$, we denote its projected location in view $v$ as $\mathbf{P}^{x,y,z}_{v}$. The heatmap feature at $\mathbf{P}^{x,y,z}_{v}$ is denoted as $\mathbf{H}_v^{x,y,z} \in \mathcal{R}^J$. We compute the feature vector of the voxel as: $\mathbf{V}^{x,y,z} = \frac{1}{V} \sum_{v=1}^{V}{\mathbf{H}_v^{x,y,z}}$ where $V$ is the number of camera views. We can see that $\mathbf{V}^{x,y,z}$ actually encodes the likelihood that the body joints are at $\mathbf{D}^{x,y,z}$. Note that the feature volume $\mathbf{V}$ is usually very noisy because some voxels which do not correspond to body joints may also get non-zero features due to lack of depth information.

We present \textbf{Joint Estimation Network (JEN)} to estimate $3$D joint heatmaps $\mathbf{U}  \in \mathcal{R}^{J,X,Y,Z}$ from $\mathbf{V}$. The network structure is shown in Figure \ref{fig:jointnetwork}. Each confidence score $\mathbf{U}^{j,x,y,z}$ represents the likelihood that there is a joint of type $j$ at voxel $\mathbf{D}^{x,y,z}$. The likelihood of all joints at all voxels form $3$D joint heatmaps $\mathbf{U}  \in \mathcal{R}^{J,X,Y,Z}$. During training, we compute ground-truth joint heatmaps $\mathbf{U}_{*}^{j,x,y,z}$ for every voxel according to its distance to ground-truth joint locations. Specifically, for each pair of ground-truth joint location and voxel, we compute a Gaussian score according to their distance. The score decreases exponentially when distance increases. Note that there could be multiple scores for one voxel if there are multiple people in the environment and we simply keep the largest one. We train JEN by minimizing:
\begin{equation}
    \mathcal{L_{\text{JEN}}}=\sum_{j=1}^{J} \sum_{x=1}^{X} \sum_{y=1}^{Y} \sum_{z=1}^{Z} {\|\mathbf{U}_{*}^{j,x,y,z} - \mathbf{U}^{j,x,y,z}\|_2}
\end{equation}

Inspired by the \emph{voxel-to-voxel} prediction network in \cite{moon2018v2v}, we adopt $3$D convolutions as the basic building block for estimating $3$D joint heatmaps. Since input feature volumes $\mathbf{V}$ are usually sparse and have clear semantic meanings, we propose a simpler structure than \cite{moon2018v2v} as shown in Figure \ref{fig:jointnetwork}. In some scenarios such as football court, the motion capture space can be very large which will inevitably result in high dimensional feature volumes. This will notably decrease the inference speed. We solve the problem by using sparse $3$D convolutions \cite{yan2018second} because in general the feature volume only has a small number of non-zero values. This significantly improves the inference speed in our experiments, especially for large voxel sizes. 

\subsection{$3$D Joint Grouping}
Suppose we have already estimated $3$D joint locations (represented by $3$D joint heatmaps) as in the above section. The remaining task is to assemble the estimated joints into poses of different instances. To that end, we present an \emph{Ambiguity Resolution Network (ARN)} to fulfill the task which is shown in the bottom part of Figure \ref{fig:jointnetwork}. We first obtain a number of $3$D pelvis joint locations by finding peak responses in the $3$D joint heatmaps. Each location represents a candidate instance of person. We perform Non-Maximum Suppression (NMS) based on the heatmap scores to extract local peaks. Then for each person, we pool features around the pelvis joint from the $3$D joint heatmaps with a fixed size $X' \times Y' \times Z'$ which is sufficiently large to enclose a person in arbitrary poses. We set $X' = Y' = Z' = 32$, which corresponds to 2000mm in real world. We feed the pooled features to ARN to estimate a $3$D pose heatmap $\mathbf{A}_k \in \mathcal{R}^{X',Y',Z'}$ for each joint $k$ of this person. The joint responses of other persons are learned to be suppressed by ARN. Finally, we use a soft argmax operation \cite{iskakov2019learnable} to generate $3$D location $\mathbf{J}_k$ of the joint $k$ from the pose heatmaps. It can be obtained by computing the center of mass of $\mathbf{A}_k$ according to the following formula:
\begin{equation}
\label{formula:integration}
    \mathbf{J}_k = \sum_{x=1}^{X'}\sum_{y=1}^{Y'}\sum_{z=1}^{Z'} (x, y, z) \cdot {\mathbf{A}_k}(x,y,z)
\end{equation}
Note that we do not obtain the location $\mathbf{J}_k$ by finding the maximum of $\mathbf{A}_k$ because the quantization error of 62.5mm is still large. Computing the expectation as in the above equation effectively reduces the error. We train ARN by comparing the estimated $3$D poses to the ground-truth $3$D poses $\mathbf{J}_*$ with $L_1$ loss as follows:
\begin{equation}
    \mathcal{L_{\text{ARN}}}=\sum_{k=1}^{J} {\|\mathbf{J}_{*}^{k} - \mathbf{J}^{k}\|_1}
\end{equation}

%% file: 3voxeltrack2.tex
We now present the second part of \emph{VoxelTrack} which links the estimated $3$D poses over time. This is a standalone module which does not require training. The core is to compute a similarity matrix between $3$D poses of the current and subsequent frame. With the similarity matrix, it accomplishes linking by solving a standard linear bipartite matching problem.

\subsection{Occlusion Relationship Reasoning}
\label{sec:occlusion_m}
Since Re-ID features in our work are extracted from images, they suffer from occlusion. Considering that most occlusion in the benchmark datasets belongs to person-person occlusion, we estimate to what extent a $3$D pose is occluded by other people in the environment. The idea can also be used to handle human-object occlusion. 

\begin{figure}
	\centering
	\includegraphics[width=3.5in]{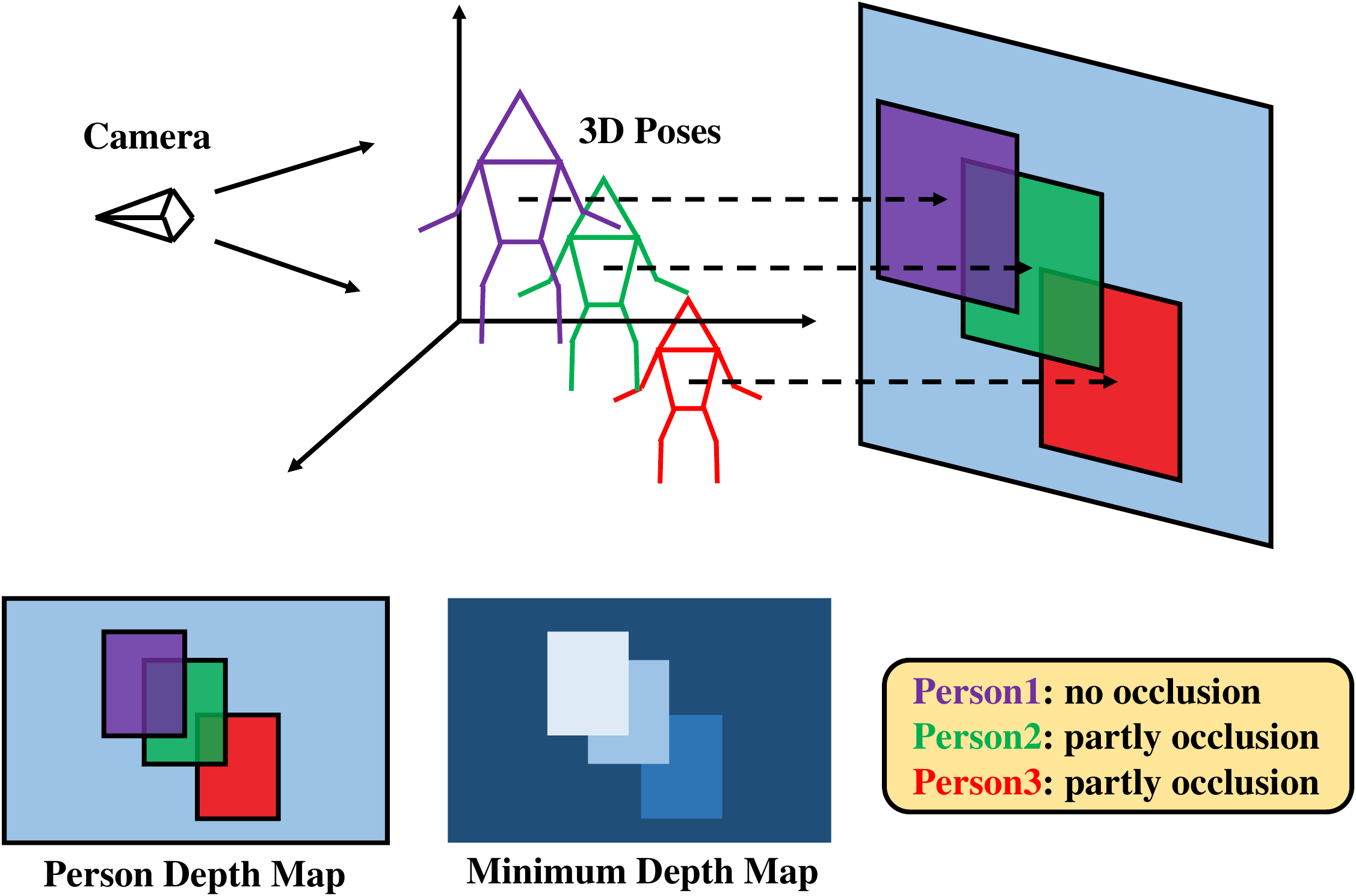}
	\caption{Some steps to compute the occlusion relationship based on depth. The person depth map is computed using camera parameters and $3$D poses. For the minimum depth map, the color becomes deeper as the depth becomes larger. 
	}
	\label{fig:occlusion}
\end{figure}

In general, if a person is severely occluded by other people in one view, we decrease the contribution of the corresponding Re-ID feature. To achieve the target, for each estimated $3$D pose, we estimate its approximate depth relative to each camera using the camera parameters. Figure \ref{fig:occlusion} shows the way to compute how much a person is occluded. Specifically, we use the average depth of all joints to represent the depth of a person. For every camera, we put a $2$D bounding box parallel to the camera plane at the average depth tightly enclosing all body joints. For each location in the box, we can easily determine whether it is occluded by the boxes of other poses. We first compute a minimum depth map for each pixel of the image. Then, we can compute the occluded area by comparing each person's depth map to the minimum depth map. The percentage of locations that are not occluded is used as a reliability score for its Re-ID feature.

\subsection{Similarity Metrics}
We compute similarity between two $3$D poses according to their appearance and spatial features. We project the $3$D pelvis joint locations of the $3$D poses to all cameras and sample corresponding Re-ID features. Denote the Re-ID features of the $i_{\text{th}}$ and $j_{\text{th}}$ poses in all camera views as $\{\mathbf{G}_i^1, \mathbf{G}_i^2, \cdots, \mathbf{G}_i^V\}$ and $\{\mathbf{G}_j^1, \mathbf{G}_j^2, \cdots, \mathbf{G}_j^V\}$, respectively. We compute the weighted average of all camera views as the final Re-ID feature $\mathbf{G}_i = \sum_{v=1}^V \omega_i^v {\mathbf{G}_i^v}$ where $\omega_i^v$ represents the reliability score computed according to occlusion relationship. In particular, if more than $70\%$ of the box is occluded by other people, $\omega_i^v$ is set to be $0$. We compute the cosine distance between the fused Re-ID features as the appearance cues. We also compute Euclidean distance between two poses to promote smoothness of tracking as the location cues. For each tracklet, we normalize the Euclidean distance between it and all the detections. We compute the average of the appearance and location similarity as the final metric.

\subsection{Tracking Framework}
We adopt a simple framework for online multiple object tracking. In the first frame, we initialize the estimated $3$D poses as tracklets. We use the Hungarian algorithm to assign the $3$D poses in the current frame to the existing tracklets. We prevent a $3$D pose from being matched to a tracklet which has very large distance. If the spatial distance between the tracklet and the $3$D pose is too large, we reject the assignment. If a $3$D pose is not matched to any tracklets, we start a new track one. When a tracklet is not matched to any $3$D poses for more than $30$ frames, we set the tracklet to inactive state and it will not be used in the future. We use the appearance features of the newly matched $3$D pose to update the appearance features of the tracklet by linear blending following \cite{zhang2020fairmot}.

%% file: 5datasets.tex
\subsection{Datasets}
\textbf{Campus Dataset \cite{belagiannis20143d}}
It captures three people interacting with each other in an outdoor environment by three cameras. We follow \cite{belagiannis20143d,dong2019fast} to split the dataset into training and testing subsets. To avoid over-fitting to this small training data, we train the $2$D part on the COCO Keypoint dataset \cite{lin2014microsoft} and train the $3$D part using the camera parameters of the Campus dataset to generate the sythetic $3$D poses and $2$D heatmap pairs. 
\vspace{0.5em}

\noindent
\textbf{Shelf Dataset \cite{belagiannis20143d}}
It captures four people disassembling a shelf by five cameras. This dataset has more occlusion than the Campus dataset. Similar to what we do for Campus, we do not use the images or poses to train on the Shelf dataset. 
\vspace{0.5em}

\noindent
\textbf{CMU Panoptic Dataset \cite{Joo_2017_TPAMI}
}
This is a recently introduced large scale multi-camera dataset for $3$D pose estimation and tracking. It captures people doing daily activities by dozens of cameras among which five HD cameras (3, 6, 12, 13, 23) are used in our experiments. We also report results when we use even fewer cameras. Following \cite{xiang2019monocular}, the training set consists of the following sequences: \seqsplit{``160422\_ultimatum1'', ``160224\_haggling1'', ``160226\_haggling1'', ``161202\_haggling1'', ``160906\_ian1'', ``160906\_ian2'', ``160906\_ian3'', ``160906\_band1'', ``160906\_band2'', ``160906\_band3''.} The testing set consists of :\seqsplit{ ``160906\_pizza1'', ``160422\_haggling1'', ``160906\_ian5'', and ``160906\_band4''}.
\vspace{0.5em}

\subsection{Metrics}
\label{sec:metrics}
\noindent
\textbf{$3$D Pose Estimation Metric} Following \cite{dong2019fast}, we use the Percentage of Correct Parts (PCP3D) metric to evaluate the estimated $3$D poses. Specifically, for each ground-truth $3$D pose, it finds the closest pose estimate and computes percentage of correct parts. We can see that this metric does not penalize false positive pose estimates. To overcome the limitation, we also extend the Average Precision (AP$_{K}$) metric \cite{pishchulin2016deepcut} in object detection to evaluate multi-person $3$D pose estimation quality which is more comprehensive than PCP3D. In particular, if Mean
Per Joint Position Error (MPJPE) of an estimate is smaller than $K$ millimeters, we think the pose is accurately estimated. AP$_{K}$ computes the average precision value for recall value over $0$ and $1$. 

\vspace{1em}

\noindent
\textbf{$3$D Pose Tracking Metric}
We modify the standard bounding box Multi-Object Tracking (MOT) metric \emph{CLEAR} \cite{bernardin2008evaluating} and the $2$D pose tracking metrics \cite{iqbal2017posetrack} for $3$D pose tracking. In particular, we compute a MOTA score for each body joint independently in a similar way as $2$D pose tracking. In particular, the matching threshold of the predicted joint and the ground truth joint is half of a head size (150mm). The MOTA score jointly considers the pose estimation and the pose linking accuracy. We count the identity switches (ID Switch) for each joint. We also compute IDF1 scores \cite{ristani2016performance} to make an overall evaluation of the identification task.

%% file: 6experiments.tex
\subsection{Implementation Details}
The training and testing images are resized to $800 \times 608$. The resulting $2$D heatmaps and Re-ID feature maps have the resolution of $200 \times 152$. We use DLA-34 \cite{zhou2019objects} as our backbone which is pre-trained on the ImageNet classification dataset. The number of body joints $J$ is set to be $15$ in accordance with the COCO dataset. The dimension of Re-ID features is set to be $64$ following FairMOT \cite{zhang2020simple}.

The motion capture space is set to be $10\text{m} \times 10\text{m} \times 4\text{m}$ for the three datasets. We divide the space into $160 \times 160 \times 64$ bins. So each bin is approximately of size $62.5\text{mm} \times 62.5\text{mm} \times 62.5\text{mm} $. Note that since we compute expectation of joint locations over $3$D heatmaps, the actual error is much smaller than $62.5$mm. Recall that we apply sparse $3$D convolution to feature volume to estimate $3$D heatmaps. To promote sparsity, we set features in the volume to zero if their original values is smaller than $0.15$. For ARN, we pool features from the space of size $2000\text{mm} \times 2000\text{mm} \times 2000\text{mm}$ (about $32 \times 32 \times 32$ voxels) around estimated pelvis joints.

We train \emph{VoxelTrack} in three separate stages. We use Adam optimizer in all stages. In the first stage, we train the $2$D model for estimating heatmaps and Re-ID feature maps for $20$ epochs with a start learning rate of $1e^{-4}$. The learning rate decreases to $1e^{-5}$ after the $15_{\text{th}}$ epoch. Next, we fix the $2$D model and train the $3$D joint estimation network for $10$ epochs. The learning rate is set to be $1e^{-4}$. Finally, we train ARN to estimate $3$D poses of all instances for $10$ epochs with learning rate set to be $1e^{-4}$. Note that we can also jointly train the three models if we have access to a large number of paired (image, $3$D pose) training data. There are several reasons why we choose separate training: (1) when we apply our model to a new environment, it may be impossible to obtain a large number of (image, $3$D pose) pairs for training the model. In this case, we can train the $2$D model on public datasets and train the $3$D model by generating a large number of $3$D pose and $2$D heatmap pairs; (2) separate training allows us to use larger batch size which helps stabilize training. 

For training on the Campus and Shelf dataset \cite{belagiannis20143d}, we do not use the images or poses and only use the camera parameters to avoid over-fitting to these small datasets. We use \cite{cheng2020bottom} as our $2$D backbone network and train on the COCO Keypoint dataset \cite{lin2014microsoft}. It is worth noting that COCO does not have identity annotations and we cannot directly train the Re-ID branch on it. We propose a weakly supervised learning approach to train the Re-ID part on the COCO dataset. We assign each $2$D pose a unique identity and thus regard each object instance in the dataset as a separate class. We apply different transformations to the whole image including flipping, rotation, scaling and translation to help create different appearances of the same instance. 

For training the $3$D part on the Campus and Shelf dataset, we generate many synthetic heatmaps using the camera parameters. we place a number of 3D poses (sampled from the motion capture datasets such as Panoptic \cite{Joo_2017_TPAMI}) at random locations in the space and project them to all views to get the respective 2D locations. Then we generate 2D heatmaps from the locations
to train the $3$D part. This has significant practical values as we can easily apply our model to new environments such as a retail store with the camera parameters available.

\begin{table*}[]
    \setlength{\tabcolsep}{8pt}
    \centering
    \begin{tabular}{r|cccc|ccc}
        \toprule
         Campus &  Actor 1 & Actor 2 & Actor 3 & Average PCP$3$D & MOTA & ID Switch & IDF1\\
         \hline
         Belagiannis \etal \cite{belagiannis20143d} & 82.0 & 72.4 & 73.7 & 75.8 & - & - & -\\
         Belagiannis \etal \cite{belagiannis2014multiple} & 83.0 & 73.0 & 78.0 & 78.0 & - & - & -\\
         Belagiannis \etal \cite{belagiannis20153d} & 93.5 & 75.7 & 84.4 & 84.5 & - & - & -\\
         Ershadi-Nasab \etal \cite{ershadi2018multiple} & 94.2 & 92.9 & 84.6 & 90.6 & - & - & -\\
         Dong \etal \cite{dong2019fast} & 97.6 & 93.3 & 98.0 & 96.3 & - & - & -\\
         Chen \etal \cite{chen2020cross} & 97.1 & \textbf{94.1} & \textbf{98.6} & 96.6 & - & - & -\\
         Ours & \textbf{98.1} & 93.7 & 98.3 & \textbf{96.7} & 89.3 & 0 & 94.6\\
         \hline
         \hline
          Shelf & Actor 1 & Actor 2 & Actor 3 & Average & MOTA & ID Switch & IDF1\\
          \hline
          Belagiannis \etal \cite{belagiannis20143d} & 66.1 & 65.0 & 83.2 & 71.4 & - & - & -\\
          Belagiannis \etal \cite{belagiannis2014multiple} & 75.0 & 67.0 & 86.0 & 76.0 & - & - & -\\
          Belagiannis \etal \cite{belagiannis20153d} & 75.3 & 69.7 & 87.6 & 77.5 & - & - & -\\
          Ershadi-Nasab \etal \cite{ershadi2018multiple} & 93.3 & 75.9 & 94.8 & 88.0 & - & - & -\\
         Dong \etal \cite{dong2019fast} & 98.8 & 94.1 & \textbf{97.8} & 96.9 & - & - & -\\
         Chen \etal \cite{chen2020cross} & \textbf{99.6} & 93.2 & 97.5 & 96.8 & - & - & -\\
         Ours & 98.6 & \textbf{94.9} & 97.7 & \textbf{97.1} & 94.4 & 0 & 97.2\\
         \bottomrule
    \end{tabular}
    \caption{Comparison to the state-of-the-art methods on the Campus and the Shelf datasets. The metric is PCP3D and AP. }
    \label{tab:campus_and_shelf}
\end{table*}

\subsection{Comparison to the State-of-the-art Methods}
\textbf{$3$D Pose Estimation } Table \ref{tab:campus_and_shelf} shows the $3$D pose estimation results of the state-of-the-art methods on the Campus and the Shelf datasets in the top and bottom sections, respectively. We can see that our approach improves PCP3D from $96.6\%$ of \cite{chen2020cross} to $96.7\%$ on the Campus dataset and $96.9\%$ of \cite{dong2019fast} to $97.1\%$ on the Shelf dataset, which is a decent improvement considering the already very high accuracy. As discussed in Section \ref{sec:metrics}, the PCP3D metric does not penalize false positive estimates. However, it is also meaningless to report AP scores because the GT pose annotations in this dataset are incomplete. So we propose to visualize and publish all of our estimated poses of the Shelf dataset\footnote{https://youtu.be/hwdk3sQEdY8} and the Campus datset\footnote{https://youtu.be/mXSQLDh953E}. We find that our approach usually gets accurate estimates as long as joints are visible in at least two views. The previous works \cite{belagiannis20143d,belagiannis20153d,belagiannis2014multiple,dong2019fast,chen2020cross} did not report numerical results on the large scale Panoptic dataset. We encourage future works to do so as in Table \ref{tab:panoptic_3d} and Table \ref{tab:panoptic_2d}.

\begin{figure*}
	\centering
	\includegraphics[width=7in]{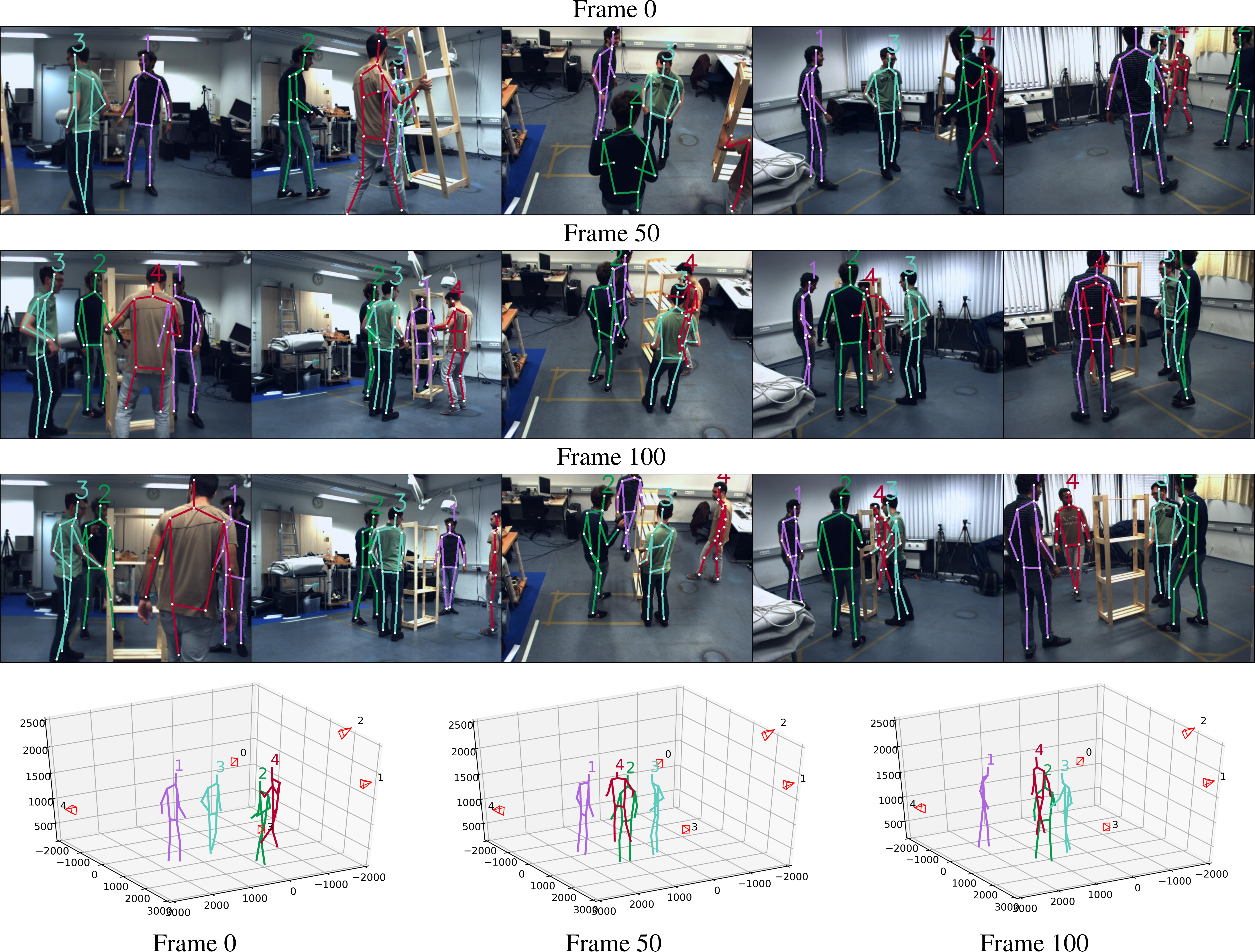}
	\caption{Visualization results on the Shelf dataset. The top is the $2$D images captured by 5 cameras and the bottom is the $3$D pose tracking results. Different numbers and colors represent different person identities. 
	}
	\label{fig:shelf}
\end{figure*}

\vspace{1em}
\noindent
\textbf{$3$D Pose Tracking } Table \ref{tab:campus_and_shelf} shows the $3$D pose tracking results of our method on the Campus and the Shelf datasets. The MOTA metric is computed by FP, FN and ID Switch, which jointly consider the person detection, pose estimation and pose tracking performance. The ID Switch and IDF1 metrics can better reveal the tracking performance. We achieve 0 ID Switch and high IDF1 score (94.6 on Campus and 97.2 on Shelf) on both of the datasets with severe occlusion in each single view, which indicates that our multi-view $3$D tracking method can achieve accurate tracking results. We achieve higher MOTA on the Shelf dataset than on the Campus dataset (94.4 vs 89.3) mainly because the pose estimation results on the Shelf dataset is better.

\vspace{1em}
\noindent
\textbf{Qualitative Study } We show some $3$D pose tracking results on the Shelf dataset in Figure \ref{fig:shelf}. We can see that there is severe occlusion in the images of all camera views. However, by fusing heatmaps from multiple cameras, our approach obtains more robust features which allows us to successfully estimate the 3D poses without bells and whistles. It is noteworthy that we do not need to associate 2D poses in different views based on noisy 2D poses by combining a number of sophisticated techniques. This significantly improves the robustness of the approach. We can see that people with identity 2, 3 and 4 are walking around the shelf with lots of occlusion. The identities of the four people keep the same across the frames, which indicates that our approach has stable tracking performance in severe occlusion cases. 

\subsection{Factors that Impact Estimation Accuracy}
We conduct ablation studies to evaluate a variety of factors of our approach. The results on the Panoptic dataset \cite{Joo_2017_TPAMI} are shown in Table \ref{tab:panoptic_3d} and Table \ref{tab:panoptic_2d}. We evaluate both the $3$D pose estimation accuracy, the $3$D tracking accuracy and the computation time. Table \ref{tab:panoptic_3d} shows the $3$D factors and Table \ref{tab:panoptic_2d} shows the $2$D factors.

\vspace{0.5em}
\noindent
\textbf{Voxel Size }
We evaluate three different voxel sizes: $160 \times 160 \times 64$, $120 \times 120 \times 48$ and $80 \times 80 \times 32$. The motion capture space is set to be $10\text{m} \times 10\text{m} \times 4\text{m}$. By comparing the first three lines in Table \ref{tab:panoptic_3d}, we can see that increasing the voxel size from $80 \times 80 \times 32$ to $120 \times 120 \times 48$ significantly improves the $3$D pose estimation accuracy as the AP$_{25}$ metric improves from 39.74 to 71.00 and the MPJPE metric decreases from 26.16mm to 19.83mm. When further increasing the size from $120 \times 120 \times 48$ to $160 \times 160 \times 64$, the improvement is not that large, which is also reasonable because the accuracy is more difficult to be increased as the grids become more fine-grained. By comparing the MOTA, IDF1 and ID Switch metrics, we can see that the voxel size does not influence the tracking accuracy much. The computation time of JEN increases as the voxel size increases. To strike a good balance between accuracy and speed, we use $160 \times 160 \times 64$ for the rest of the experiments.

\begin{table*}[]
    \setlength{\tabcolsep}{7pt}
    \centering
    \begin{tabular}{cc|cccc|ccc|cc}
        \toprule
         Voxel Size & JEN Type & $AP_{25}$ & $AP_{50}$ & $AP_{100}$ & MPJPE & MOTA & IDF1 & ID Switch & JEN Time & ARN Time\\
         \midrule
         $160 \times 160 \times 64$ & SP Conv & 79.34 & 96.83 & 99.58 & 18.49 mm & 98.45 & 98.67 & 0 & 48.46 ms & $2.72 \times n$ ms\\
         $120 \times 120 \times 48$ & SP Conv & 71.00 & 97.04 & 99.45 & 19.83 mm & 98.27 & 98.52 & 0 & 30.93 ms & $1.35 \times n$ ms\\
         $80 \times 80 \times 32$ & SP Conv & 39.74 & 94.27 & 99.10 & 26.16 mm & 97.62 & 95.13 & 0 & 22.01 ms & $1.21 \times n$ ms\\
         $160 \times 160 \times 64$ & Conv & 74.09 & 96.87 & 99.55 & 19.05 mm & 98.32 & 98.39 & 0 & 132.64 ms & $2.71 \times n$ ms\\
         $120 \times 120 \times 48$ & Conv & 68.89 & 97.06 & 99.51 & 20.28 mm & 98.16 & 98.21 & 0 & 57.30 ms & $1.36 \times n$ ms\\
         $80 \times 80 \times 32$ & Conv & 38.66 & 94.40 & 99.17 & 25.93 mm & 98.27 & 98.56 & 0 & 18.75 ms & $1.21 \times n$ ms\\
         \bottomrule
    \end{tabular}
    \caption{Ablation study of voxel size and sparse convolution on the Panoptic dataset. }
    \label{tab:panoptic_3d}
\end{table*}

\vspace{0.5em}
\noindent
\textbf{Sparse Convolution} We use the sparse convolution to replace the standard convolution as the $3$D feature volume only has a small number of non-zero values and the sparse convolution only computes for the non-zero values. By comparing the sparse convolution to the standard convolution with the same voxel size in Table \ref{tab:panoptic_3d}, we can see that the computation time of JEN significantly decreases when using sparse convolution, especially under large voxel sizes such as $160 \times 160 \times 64$ and $120 \times 120 \times 48$. For the small voxel size such as $80 \times 80 \times 32$, the sparse convolution is a little slower than the standard convolution. This is because the sparse convolution needs to find the index of the non-zero values and it takes considerable time. Thus, we do not apply sparse convolution to ARN because the size of the input $3$D heatmap to ARN is much smaller. We can also see that the pose estimation accuracy of the sparse convolution is a little higher than the standard convolution under all voxel sizes. This is because we set the values of the feature volume to zero if their original values is smaller than 0.15 when applying the sparse convolution. The sparsity of the feature volume may reduce some ambiguity and thus increase the pose estimation accuracy.

\begin{table*}[]
    \setlength{\tabcolsep}{8pt}
    \centering
    \begin{tabular}{ccc|cccc|ccc|c}
        \toprule
         Views & Backbone & Image Size & $AP_{25}$ & $AP_{50}$ & $AP_{100}$ & MPJPE & MOTA & IDF1 & ID Switch & $2$D Time\\
         \midrule
         5 & DLA-34 & $960 \times 512$ & 79.34 & 96.83 & 99.58 & 18.49 mm & 98.45 & 98.67 & 0 & 85.71 ms\\
         4 & DLA-34 & $960 \times 512$ & 66.20 & 96.34 & 99.47 & 20.35 mm & 98.37 & 98.46 & 0 & 66.93 ms\\
         3 & DLA-34 & $960 \times 512$ & 49.09 & 92.44 & 97.62 & 24.93 mm & 95.77 & 93.08 & 0 & 54.99 ms\\
         5 & DLA-34 & $800 \times 448$ & 70.66 & 97.26 & 99.70 & 19.99 mm & 98.61 & 98.99 & 0 & 65.20 ms\\
         5 & DLA-34 & $640 \times 384$ & 55.96 & 96.78 & 99.65 & 21.67 mm & 98.37 & 98.45 & 0 & 45.37 ms\\
         5 & MobileNet-V2 & $960 \times 512$ & 42.42 & 94.09 & 99.33 & 24.38 mm & 97.61 & 97.82 & 0 & 27.50 ms\\
         5 & Higher-HRNet-W32 & $960 \times 512$ & 85.88 & 98.31 & 99.54 & 16.97 mm & 98.51 & 98.73 & 0 & 128.95 ms\\
         \bottomrule
    \end{tabular}
    \caption{Ablation study of number of views, 2D backbone and input image size on the Panoptic dataset. }
    \label{tab:panoptic_2d}
\end{table*}

\begin{table*}[]
    \setlength{\tabcolsep}{9pt}
    \centering
    \begin{tabular}{ccc|ccc|c}
        \toprule
         Re-ID Features & Occlusion Mask & 3D Poses & MOTA & IDF1 & ID Switch & Tracking Time\\
         \midrule
          &  & $\surd$ & 98.42 & 93.82 & 90 & 0.92 ms\\
         $\surd$ &  &  & 98.44 & 94.38 & 15 & 0.96 ms\\
         $\surd$ & $\surd$ &  & 98.45 & 98.67 & 0 & 2.10 ms\\
         $\surd$ & $\surd$ & $\surd$ & 98.45 & 98.67 & 0 & 2.16 ms\\
         \bottomrule
    \end{tabular}
    \caption{Ablation study of Re-ID features, occlusion mask and 3D poses on the Panoptic dataset. }
    \label{tab:panoptic_tracking}
\end{table*}

\vspace{0.5em}
\noindent
\textbf{Number of Cameras }
As shown in the first three lines of Table \ref{tab:panoptic_2d}, reducing the number of cameras generally increases the $3$D pose estimation error because the information in the feature volume becomes less complete. The tracking accuracy is less affected by the number of cameras as the ID Switch is always 0. We can see that using 3 cameras can already accurately track the $3$D poses. The computation time also decreases as the number of cameras decreases. 

\vspace{0.5em}
\noindent 
\textbf{Backbone Networks}
We evaluate three different backbone networks including DLA-34 \cite{zhou2019objects}, MobileNet-V2 \cite{sandler2018mobilenetv2} and Higher-HRNet-W32 \cite{cheng2020bottom}. For the MobileNet-V2, we add several de-convolution layers after the backbone network following \cite{xiao2018simple}. The results are shown in Table \ref{tab:panoptic_2d}. Higher-HRNet-W32 achieves the highest AP and the lowest MPJPE. MobileNet-V2 achieves the highest running speed. DLA-34 achieves a good balance between accuracy and speed. 

\vspace{0.5em}
\noindent
\textbf{Image Sizes}
We also evaluate three different image sizes including $960 \times 512$, $800 \times 448$, $640 \times 384$. As shown in Table \ref{tab:panoptic_2d}, we use DLA-34 as the backbone and evaluate different image sizes. Reducing the image sizes generally increases the $3$D pose estimation error because large size images provide more detailed information. The tracking performance is hardly affected by image sizes. We can see that getting accurate $2$D heatmaps is critical to the $3$D accuracy.

\begin{figure*}
	\centering
	\includegraphics[width=7in]{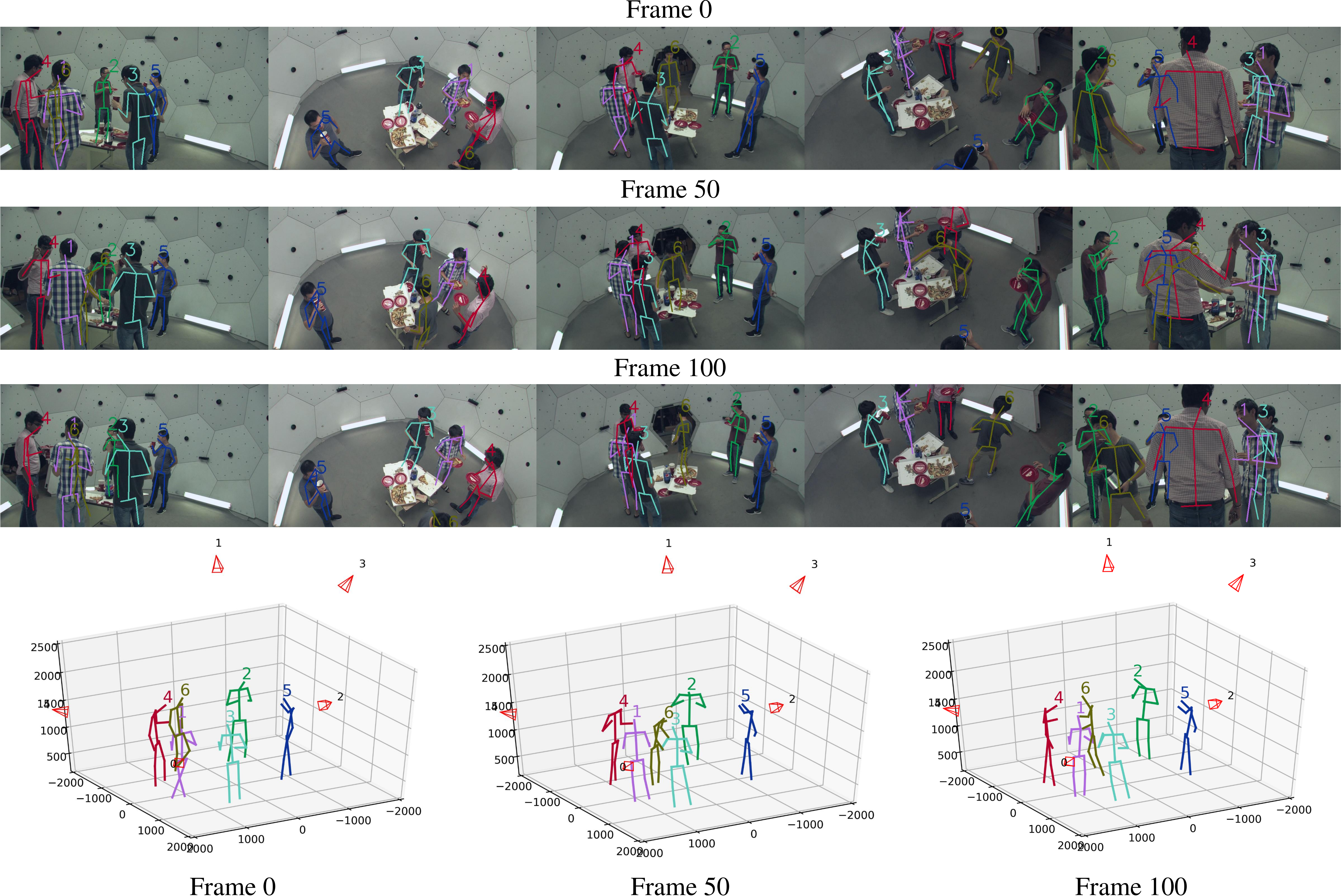}
	\caption{Visualization results on the ``160906\_pizza1'' sequence of the Panoptic dataset. The top is the $2$D images captured by 5 cameras and the bottom is the $3$D pose tracking results. Different numbers and colors represent different person identities. 
	}
	\label{fig:panoptic}
\end{figure*}

\subsection{Factors that Impact Tracking Accuracy }
There are three main components in our tracking procedure: 1) 3D poses, 2) Re-ID features, 3) occlusion-mask. We evaluate the impact of each of these components. The tracking results are shown in Table \ref{tab:panoptic_tracking}. The MOTA and IDF1 metrics are the average of all keypoints. The ID Switch metric is the switches of all the keypoints and it is a multiple of the number of joints (\ie 15). 

\vspace{0.5em}
\noindent
\textbf{$3$D Poses }
We only use the normalized Euclidean distances of the $3$D poses to link the detections to the tracklets. The result is shown in the first line of Table \ref{tab:panoptic_tracking}. We can only get 93.82 IDF1 score and 90 ID switches when only using the $3$D poses. We find that the ID switches often occur when FP appears. In general, the $3$D poses are reliable because there is almost no occlusion in the $3$D space.   

\vspace{0.5em}
\noindent
\textbf{Re-ID Features }
We only use the cosine distance of Re-ID features to perform linking. For each $3$D person, we fuse the Re-ID features in each view by just adding them without reasoning about the occlusion. The result is shown in the second line of Table \ref{tab:panoptic_tracking}. The IDF1 score (94.38 vs 93.82) and ID Switch (15 vs 90) are better than only using $3$D poses. There still exists some ID Switches because some views with heavily occluded people provide some unreliable Re-ID features which cause some ambiguity.

\vspace{0.5em}
\noindent
\textbf{Occlusion Mask}
We use the occlusion mask mentioned in Section \ref{sec:occlusion_m} computed by each person's depth to fuse the Re-ID features of all the views. If the person in the view is heavily occluded, we do not use the Re-ID features of the person in this view. The result is shown in the third line of Table \ref{tab:panoptic_tracking}. We can see that using the occlusion mask to fuse Re-ID features achieves the highest IDF1 score 98.67 and does not have ID Switch which agrees with our expectation. When we further use the Re-ID features with occlusion mask and the $3$D poses together, the tracking results keep the same, which indicates that our multi-view fused Re-ID features have powerful discriminative ability.  

\vspace{0.5em}
\noindent
\textbf{Qualitative Study }
We show the $3$D pose tracking results of the Panoptic dataset in Figure \ref{fig:panoptic}. We can see that there are severe occlusions in the images of all camera views. The person with identity 6 has the most obvious movement. He comes to the table and then leaves. He is occluded in most of the camera views and the Re-ID features of the person are not reliable in most views. Thus, we need to use the occlusion mask to choose the Re-ID features in the specific views where the person is not occluded. Our approach can keep the identities of the six people the same and has stable tracking performance.

\subsection{Whole System Running Time }
We divide our whole system into the $2$D part, the $3$D part and the tracking part and compute the running time of each of them. The running time of the $3$D part is shown in Table \ref{tab:panoptic_3d} which is the sum of the ``JEN Time'' and the ``ARN Time''. The running time of the $2$D part and the tracking part is shown in Table \ref{tab:panoptic_2d} and Table \ref{tab:panoptic_tracking}, respectively. From Table \ref{tab:panoptic_3d} we can see that the sparse convolution can reduce a large amount of running time of JEN. The voxel size also notably affects the running time. It is worth noting that the running time of ARN is hardly affected by the number of people (\ie 1 \emph{ms} for 1 person). From Table \ref{tab:panoptic_2d} we can see that the number of views, the backbone network, and the image size together determine the running time of the $2$D part. From Table \ref{tab:panoptic_tracking} we can see that the running time of the tracking part can almost be ignored (\ie 2 \emph{ms}), which indicates the simplicity of our tracking algorithm. The light version of our system using MobileNet-V2 \cite{sandler2018mobilenetv2} as the $2$D backbone and $120 \times 120 \times 48$ JEN with the sparse convolution can run at 15 FPS with 5 camera views as input, which dramatically enhances the practical values of our approach. 


%% file: 7conclusion.tex
\label{sec:conclusion}
We present a novel approach for multi-person $3$D pose estimation and tracking. It employs a multi-branch network to jointly estimate 3D poses and Re-ID features for all people in the environment. Different from the previous methods, it only makes hard decisions in the 3D space which allows to avoid the challenging association problems in the 2D space. In particular, noisy and incomplete information of all camera views are warped to a common 3D space to form a comprehensive feature volume which is used for 3D estimation. We also introduce an occlusion-aware matching strategy during tracking. The experimental results on the benchmark datasets validate that the approach is robust to occlusion. In addition, the $3$D part of the approach can be directly trained on synthetic data which has practical values. 